\documentclass[letterpaper, 10 pt, conference]{ieeeconf}  % Comment this line out if you need a4paper
\IEEEoverridecommandlockouts                              % This command is only needed if 
\usepackage{todonotes}
\usepackage{xcolor}

\title{\LARGE \bf
High Aspect Ratio Multi-stage Ducted Electroaerodynamic\\ Thrusters for Micro Air Vehicle Propulsion
}

\author{C. Luke Nelson$^{1}$ and Daniel S. Drew$^{2}$ % <-this % stops a space
\thanks{$^{1}$Department of Mechanical Engineering, University of Utah, Salt Lake City, UT 84112, USA}
\thanks{$^{2}$Department of Electrical and Computer Engineering, University of Utah, Salt Lake City, UT 84112, USA}
\thanks{Corresponding author: Daniel S. Drew, \tt{daniel.drew@utah.edu}}
}

\begin{document}

\maketitle
\thispagestyle{empty}
\pagestyle{empty}

%%%%%%%%%%%%%%%%%%%%%%%%%%%%%%%%%%%%%%%%%%%%%%%%%%%%%%%%%%%%%%%%%%%%%%%%%%%%%%%%
\begin{abstract}
Electroaerodynamic propulsion, where force is produced through collisions between electrostatically accelerated ions and neutral air molecules, is an attractive alternative to propeller- and flapping wing-based methods for micro air vehicle (MAV) flight due to its silent and solid-state nature. One major barrier to adoption is its limited thrust efficiency at useful disk loading levels. Ducted actuators comprising multiple serially-integrated acceleration stages are a potential solution, allowing individual stages to operate at higher efficiency while maintaining a useful total thrust, and potentially improving efficiency through various aerodynamic and fluid dynamic mechanisms. In this work, we investigate the effects of duct and emitter electrode geometries on actuator performance, then show how a combination of increasing cross-sectional aspect ratio and serial integration of multiple stages can be used to produce overall thrust densities comparable to commercial propulsors. An optimized five-stage device attains a thrust density of about 18 N/m$^2$ at a thrust efficiency of about 2 mN/W, among the highest values ever measured at this scale. We further show how this type of thruster can be integrated under the wings of a MAV-scale fixed wing platform, pointing towards future use as a distributed propulsion system.

\end{abstract}

%%%%%%%%%%%%%%%%%%%%%%%%%%%%%%%%%%%%%%%%%%%%%%%%%%%%%%%%%%%%%%%%%%%%%%%%%%%%%%%%
\section{Introduction}
\label{sec:intro}
Autonomous micro air vehicles (MAVs) are poised to transform the way that we gather information from the world around us. Whether it is in commercial, industrial, or defense settings, the ability to unobtrusively collect data using cheap, high agent count systems would be highly beneficial~\cite{floreano2015science,dorigo2020reflections}. Accordingly, academic research on MAV platforms ranging from fixed-wing gliders to bio-inspired flappers has grown steadily since the early 2000s~\cite{ward2017bibliometric}. The science of autonomy at small scales has also advanced: sub-gram electronics payloads are now sufficient for flight control~\cite{talwekar2022towards}, untethered flight of insect-scale robots has been shown with beamed power~\cite{james2018liftoff}, and groups of MAVs have been shown working together in unstructured environments~\cite{zhou2022swarm}.

\begin{figure}[t]
    \centering
    \includegraphics[width=0.95\columnwidth]{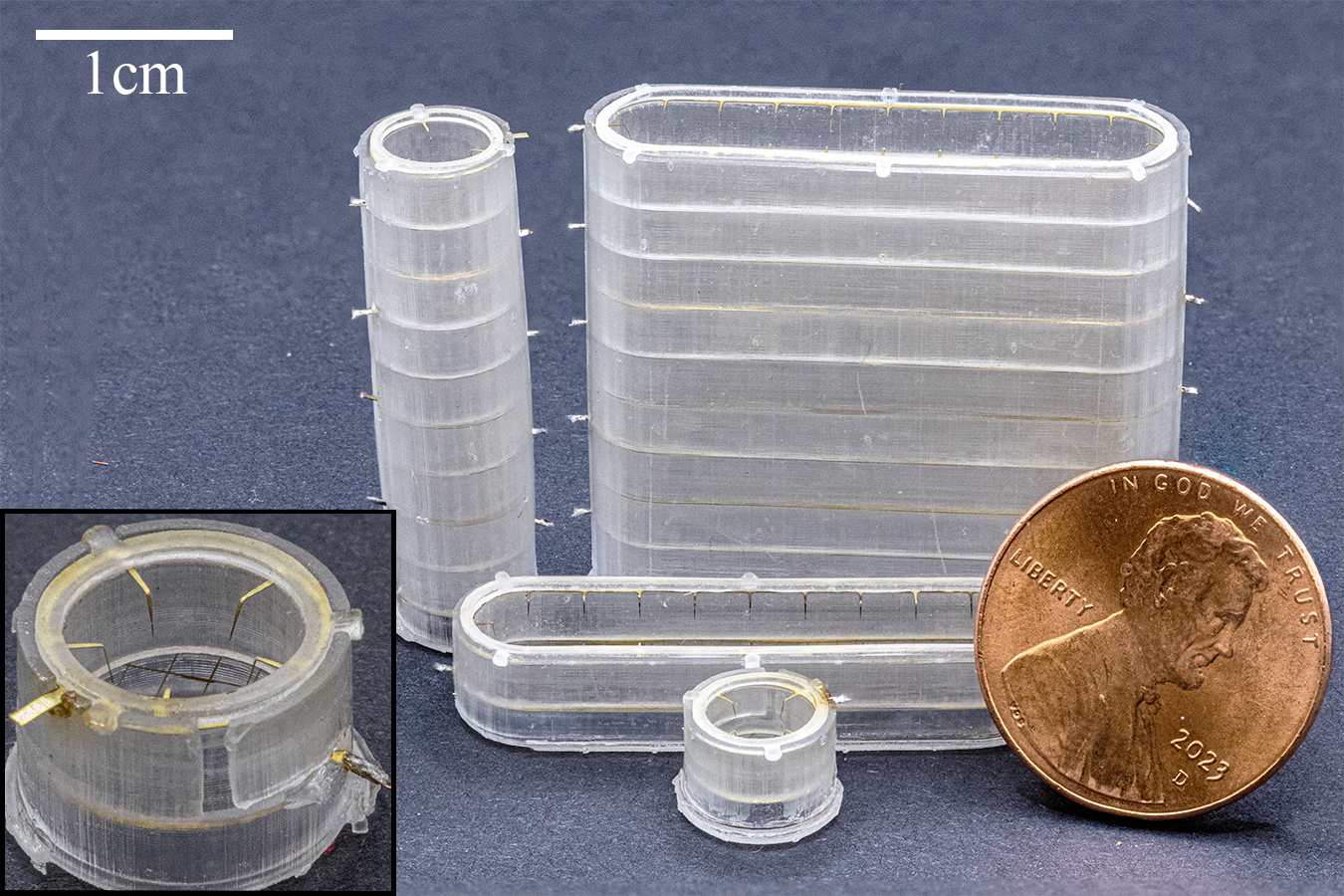}
    \vspace{-2mm}
    \caption{Multi-stage ducted electroaerodynamic thrusters next to U.S. penny for scale: a 6~mm diameter five-stage cylindrical thruster (back left), a five-stage  aspect ratio five thruster (back right), and single stage versions in front of them. Inset: Closer view at a single cylindrical stage showing the five emitter tips bent vertically down towards a collector grid.}
    \label{fig:teaser}
    \vspace{-4mm}
\end{figure}

Current commercial rotor-based MAVs, however, are not suitable for all types of envisioned deployments. Fundamental challenges like the poor aerodynamic efficiency of small propellers and the decreased performance of miniaturized electromagnetic motors means that centimeter-scale quadrotors are often limited to flight times on the order of minutes~\cite{mulgaonkar2014power}. In addition, their propellers, spinning at tens of thousands of RPM, are fragile and noisy, making them unlikely to be useful in human-proximal and space-constrained environments~\cite{schaffer2021drone}.  To overcome these challenges, researchers have turned towards biomimetic flapping-wing designs, which exhibit better aerodynamic scaling at low Reynolds numbers~\cite{wood2012progress}. Nevertheless, their demonstrated lift efficiency and achievable thrust-to-weight ratio have still yet to exceed that of a quadrotor, and the motion of their wings relies on complicated transmissions~\cite{ren_high-lift_2022}. Atmospheric ion thrusters (Fig.~\ref{fig:teaser}) are an attractive alternative, especially for fixed-wing and lighter-than-air fliers which are not easily outfitted with flapping wings. 

\begin{figure}[t]
    \centering
    \includegraphics[width=\columnwidth]{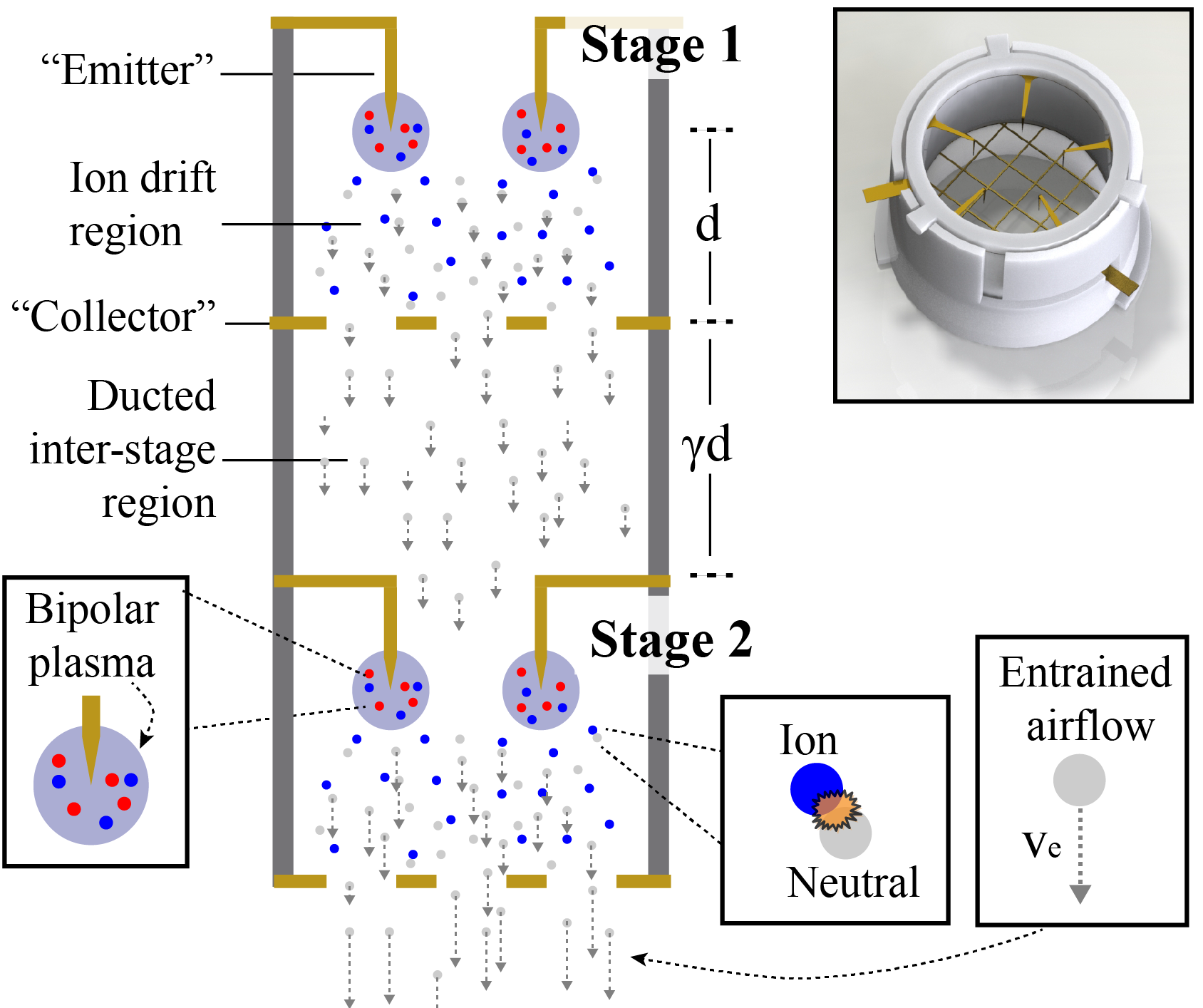}
    \caption{Schematic view of a two-stage ducted electroaerodynamic thruster in cross section. Ions are ejected from the bipolar corona plasma locally confined to the volume around the emitter tips and drift towards the collector grid under the influence of an applied electric field. Along the way, they collide with neutral air molecules and transfer their momentum, resulting in an entrained airflow through the collector grid. Critical dimensions include the inter-electrode gap distance $d$, the inter-stage distance defined as a multiple $\gamma$ of $d$, and the duct diameter $D$. }
    \label{fig:schematic}
    \vspace{-4mm}
\end{figure}

Electrohydrodynamic (EHD) propulsion, where force is produced via the momentum-transferring collisions of ions with neutral fluid molecules, works virtually silently and with no mechanical moving parts. When the impacted neutral molecules are atmospheric air, it is known as electroaerodynamic (EAD) propulsion. EAD thrusters have propelled platforms ranging from a five-meter wingspan ``solid-state aeroplane~\cite{xu2018flight}'' to the insect-scale ``ionocraft~\cite{drew2018toward}.'' It is particularly well suited for miniaturization and use in MAVs due to its cross-sectional force scaling, simple mechanical design amenable to batch manufacturing, and ability to modulate force directly with applied voltage. Corona discharge (Fig.~\ref{fig:schematic}) is the most common mechanism for ion generation in EAD actuators owing to its simplicity (e.g., a single applied potential both strikes the plasma and accelerates the ions) and for its proven reliability (e.g., it is often used as a stable ion source for spectroscopy). 

The fundamental physical phenomena governing electroaerodynamic propulsion are well understood for simple cases; at its core, performance is governed by the electric drift field magnitude accelerating the ions for their collisions with neutral air molecules~\cite{pekker2011model}. There is an inherent trade off between force and efficiency established by validated theory, with bounding values for each coarsely determined by the breakdown field in air and the space charge limit for unipolar drift current~\cite{masuyama2013performance,gilmore2015electrohydrodynamic}. A one-dimensional derivation yields equations for force, $F$, and thrust efficiency in N/W, $\eta$, as functions of electric field, ion drift current and travel distance, and ion mobility $\mu$ \cite{pekker2011model}: 

\vspace{-1em}
\begin{center}
\begin{tabular}{p{4cm}p{2cm}}
  \begin{equation}
  F = \frac{Id}{\mu} = \frac{9}{8} \epsilon_0 A E^2
  \end{equation}
  &
  \begin{equation}
  \eta = \frac{1}{\mu E}
  \end{equation} 
\end{tabular}
\vspace{-1em}
\end{center}

While these equations do not capture various loss factors specific to individual device geometries and ion generation mechanisms, it is clear that there is an inherent trade-off between thrust and thrust efficiency as dictated by the ion drift field magnitude. One method proposed to overcome this fundamental tradeoff---to produce actuators with both high thrust density and high efficiency---is to stack many small acceleration stages in series in a single ``multi-stage ducted (MSD)'' device~\cite{drew2021high,gomez2023model}. Each stage can be driven at a higher efficiency and lower thrust operating point, then multiple stages can be used to reach the total desired force output (and increase the total effective thrust density). Ducting the serially-integrated acceleration stages (Fig.~\ref{fig:schematic}) is a way to reduce losses (e.g., from collector electrode drag), safely and effectively package propulsors, and even improve overall efficiency via aerodynamic mechanisms like pressure forces acting on the duct~\cite{gomez2023model}. This multi-stage approach is especially amenable to miniaturization, as volumetric force density of an EAD device is related to its inter-electrode gap, whereas force and efficiency are governed by the scale invariant electric field magnitude. This means that in a given actuator volume, using multiple miniaturized stages instead of a single one can increase power density without affecting areal thrust or thrust efficiency. 

% The main contribution of this paper is an investigation of how various design parameters affect the propulsive performance of centimeter-scale electrohydrodynamic actuators. We include studies of laser micromachined emitter geometry for different diameter ducts, different angles of attack for cambered intake lips, different area ratios for exhaust nozzles, and how integrating successive stages (one, two, and three) affect exhaust nozzle performance. Initial numerical simulations guide our device design towards a suite of experiments enabled by an automated multi-instrument setup. Our design culminates in a three-stage device with intake lip and nozzle that exhibits over $\%$ greater thrust efficiency than a three-stage control device with only a straight duct and $\%$ greater total force relative to a single-stage control device. 

The broad objective of this work is to investigate centimeter-scale multi-stage ducted EAD thrusters and provide a feasibility proof for their use in MAV propulsion. The contributions of this paper include a study of emitter electrode and duct geometric parameters for annular ducted thrusters, demonstration of a strategy for scaling annular thrusters to high aspect ratios in order to increase force without compromising thrust density or efficiency, and characterization of multi-stage devices with more successive stages than have ever been shown at this scale. It concludes with an assembled proof-of-concept MAV-scale fixed-wing platform with underwing thrusters, pointing the way to autonomous MAVs with distributed EAD propulsion.

%a study of inter-stage region sizing for this novel electrode configuration,

%direct measurements of the glide ratio of a MAV-scale fixed wing platform during unpowered flight with and without underwing thrusters, pointing the way to autonomous fixed wings MAVs with distributed EAD propulsion.}

\section{Related Work}
\label{sec:related}
The most relevant related work concerns EAD-based fliers, ducted electrohydrodynamic pumps, and the design and use of shrouded rotors for micro air vehicles.

\subsection{EAD Propulsion for Flying Robots}
The concept of EAD-based distributed propulsion for unmanned fixed-wing flight was shown at the meter-scale by Xu et al. ~\cite{xu2018flight}, but used unducted actuators two orders of magnitude larger than in this work. Microfabricated silicon electrodes have been used to produce high ($\approx$10) thrust-to-weight ratio centimeter scale fliers~\cite{drew_first_2017}, and an EAD quad-thruster design has lifted a useful sensor payload with tethered power~\cite{drew2018toward}. Similar designs using UV laser microfabricated electrodes (as in this work) have also been shown to produce sufficient thrust for liftoff~\cite{prasad_laser-microfabricated_2020,zhang_passive_2022,zhang_centimeter-scale_2022}. None of these were multi-stage or ducted actuators.

\subsection{Ducted EAD Actuators}
While the majority of effort on EAD propulsive devices has focused on ductless designs, there is existing work studying single- and multi-stage ducted devices for the more general purpose of maximizing volumetric flow rate for fluid pumping. Kim et al. found that exhaust velocity and electromechanical efficiency increased with the square root of the number of active stages (from one to six) for a straight ducted device 78~mm in diameter~\cite{kim2010velocity}.  Moreau et al. showed that electromechanical efficiency of an EHD pump was reduced when duct size was too small relative to the inter-electrode gap, with a local maximum at some diameter between one and two times the gap~\cite{moreau2008enhancing}. Here, we investigate these trends for smaller devices at lower Reynolds numbers.

Research on ducted EAD devices specifically for propulsion is still emerging. Drew et al. found a near linear increase in output force with number of stages (from one to three) in a millimeter-scale device, but did not thoroughly investigate emitter design parameters or duct geometry~\cite{drew2021high}. Recent work by Gomez-Vega et al. \cite{gomez2023model} on an analytical model for multi-stage ducted EAD devices based on simple momentum theory shows the potential for orders of magnitude improvement in thrust at a given thrust efficiency (or vice versa) depending on inter-stage losses, though it is unclear whether their modeling assumptions hold for the low Reynolds number and more complicated electrostatic conditions found in miniaturized thrusters like in this work.

%  This approach has been recently validated at the centimeter scale empirically~\cite{drew2021high} and for the general analytical case~\cite{gomez2023model}. Existing work shows near-linear scaling of output force with number of stages \textit{n}, with loss captured by a hyperparameter $\beta$ \cite{drew2021high}; properly designed ducts have the potential to increase this loss term \textit{above} unity:

% \vspace{-1em}
% \begin{center}
% \begin{tabular}{p{3cm}p{3cm}}
%   \begin{equation}
%   F_n = n \beta F
%   \end{equation}
%   &
%   \begin{equation}
%   \textbf{f}_{s.s.} \approx F / A d  
%   \end{equation} 
% \end{tabular}
% \vspace{-1em}
% \end{center}

\subsection{Shrouded Rotors for Micro Air Vehicles}
Annular lift fans and other shrouded rotor-based propulsive systems demonstrate efficiency benefits from sources including reduced energy wasted in fluid contraction downstream of the rotor disk and reduction in tip vortices. Their use in MAVs, however, is still limited, due at least in part to challenges with thrust-to-weight ratio and meeting tight geometric tolerances at low cost. Shrouded MAV-scale rotors have been shown with thrust coefficients up to 50$\%$ higher than with an open rotor, depending on shroud expansion ratio and tip clearance \cite{pereira2008hover}, and a 6.5~cm diameter shrouded rotor has been shown to reduce power requirements for hovering by about 10$\%$ \cite{hrishikeshavan2012design}.

% Using simple momentum theory, thrust benefits for a rotor with constant input power and disc area arising due to the expansion of fluid in the shroud with exhaust area can be derived purely from expansion ratio $\sigma$ and is given as:

% \vspace{-1em}
% \begin{center}
% \begin{tabular}{p{2cm}p{4cm}}
%   \begin{equation}
%   \sigma = \frac{A_e}{A_i}
%   \end{equation}
%   &
%   \begin{equation}
%   \frac{T_{shroud}}{T_{open}} = (2\sigma)^{1/3}
%   \end{equation} 
% \end{tabular}
% \vspace{-1em}
% \end{center}

% For an open rotor, the expansion ratio is 0.5. For an expansion ratio of 1 (a straight cylindrical duct), we would therefore expect a thrust gain of ~25$\%$ at the same input power. In real experiments with rotors, performance gains are even higher than this momentum theory prediction due to decreases in losses at the rotor tips: 

% \dsd{Should include a couple of sentences here talking explicitly about what benefits the model in \cite{gomez2023model} predicts from ducting, comparing directly to the equation we just gave.} 
% In addition, \cite{gomez2023model} predicts ducting EAD devices will improve the output thrust due to pressure forces acting on the duct by taking advantage of pressure accumulation over additional stages in contrast to the atmospheric pressure at the outlet. 
%Note for Dan I'm assuming we talk about shrouded rotors as an explanation for why we are ducting the EAD devices, Correct? Yes.

\section{Methods and Approach}
\label{sec:methods}
\subsection{Electrode Design}
Emitter and collector electrodes are fabricated using a 355-nm UV laser micromachining system (DPSS Samurai UV Marking System) to etch 25~$\mu$m thick brass shim stock. Duct structures are fabricated via stereolithographic printing (Formlabs Form 3+, clear resin). 

Prior work has shown that lithographically-defined emitter asperities can reduce corona plasma onset voltage and increase subsequent ion current compared to a bare wire \cite{drew2017geometric}. Here, we define emitter tips lithographically and arrange them inwards from an annulus or rounded rectangle, depending on the device aspect ratio. Each emitter tip is drawn as a triangle with a height defined by the difference in diameters between the outer circle where the triangles begin, $d_2$, and an inner circle where their tips reach, $d_1$. Their tip-to-tip distance is given by chord length $c = d_1 sin(\pi / n$), where $n$ is the number of tips. Each emitter tip is bent downwards 1 mm at a right angle where it meets the edge of the inner circle, yielding a distance from the tip to the annulus lip of $\sqrt{(d_2-d_1)^2 + 1}$.
The field enhancement factor of the emitter tips, responsible in part for determining the corona discharge performance, results from a combination of the effective radius of curvature of the tip after etching and electrostatic shielding from the metal rim of the electrode and from neighboring emitter tips. Preliminary process development determined the minimum nominal tip angle repeatably achievable in the UV laser micromachining process to be $5^\circ$, which is used for all emitters. All devices use a 2~mm inter-electrode (emitter tip to collector grid) gap.

% \textcolor{red}{$c = d_1 sin(2\pi / n)$,} or \textcolor{red}{$c = 2(d_1radius) cos(180^\circ(n-2)/n /2)$,}

Based on prior success with point-to-grid designs and a desire to focus on emitter effects in this work, we opted for a grid collector geometry. The collector grids must balance presenting a quasi-uniform potential plane in order to maximize discharge performance with the desire to minimize aerodynamic drag on the air which is accelerated through it via ion-neutral collisions. Preliminary results found that grids with a wire width of 50~$\mu$m and a wire spacing of 1~mm had no effect on corona onset voltage compared to less sparse designs, and were able to be repeatability fabricated and assembled. This agrees with prior work, which found that a grid spacing equal to approximately half the inter-electrode distance did not influence discharge performance~\cite{drew2017geometric}.

% \subsection{Duct Design}
%  We explore this value experimentally for our unique geometry, where emitters project out and down from an annulus over a grid.

\subsection{Fabrication and Assembly}
\begin{figure}[t]
    \centering
    \includegraphics[width=\columnwidth]{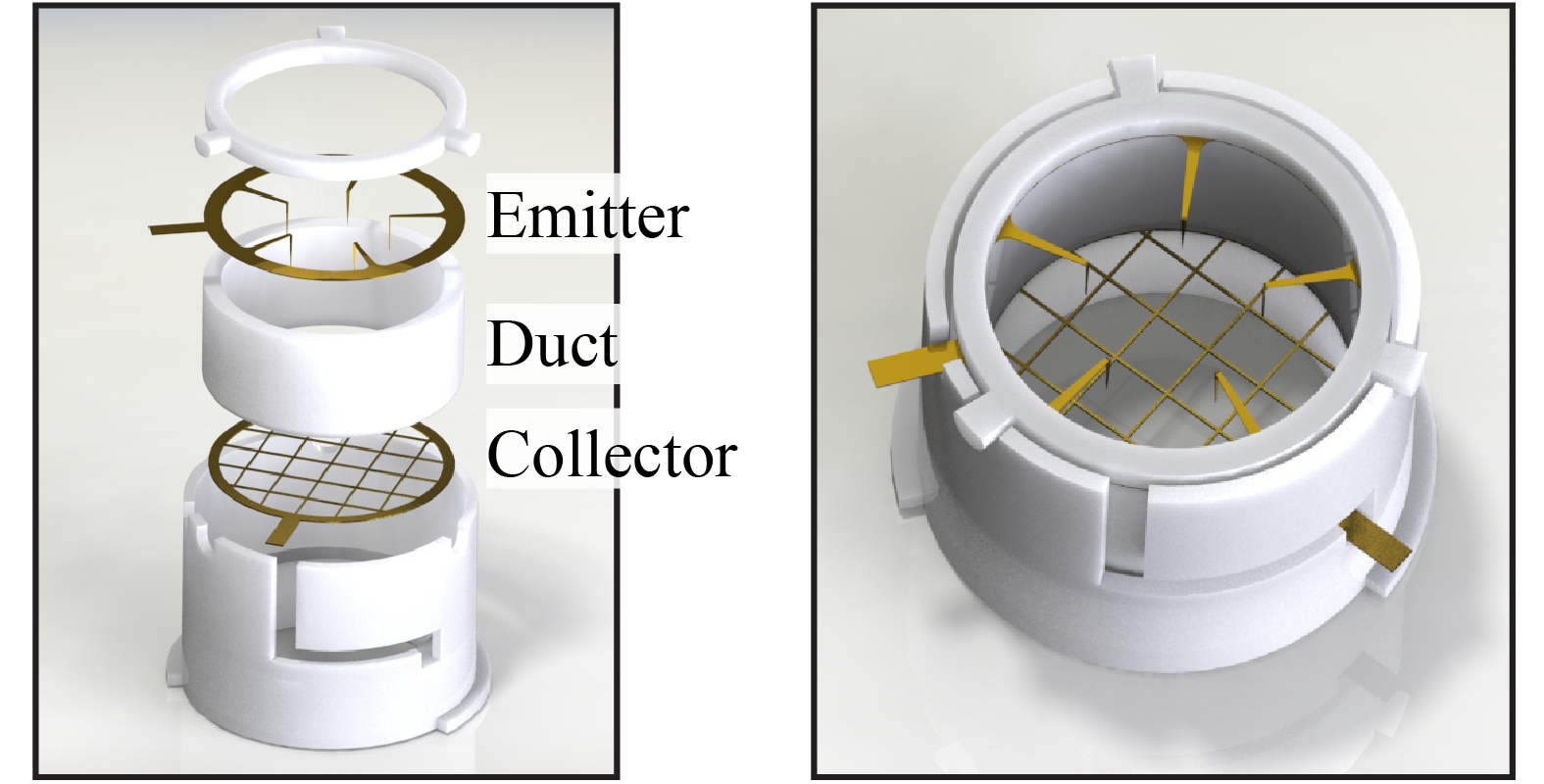}
    \caption{Rendered view of the ducted electroaerodynamic thrusters explored in this work, with individual components separated to show the assembly process. Laser micromachined active electrodes are integrated with SLA-printed duct structures in an adhesiveless process based on precise mechanical affordances. Prior to insertion in the duct, emitter electrodes are bent out of plane using an SLA-printed stamp and die set (not shown).}
    \label{fig:renders}
    \vspace{-4mm}
\end{figure}

Reliable fabrication and assembly is seen as one of the major barriers to progress in novel micro (and smaller) air vehicle development~\cite{wood2012progress}. Producing millimeter-scale EAD actuators, which still operate at high voltage, is particularly challenging given their tendency to destructively arc. Integrating multiple stages further compounds error from device variability, and has limited the impact of existing empirical work in this space~\cite{drew2021high}. We have developed a new methodology to produce modular millimeter-scale actuators based on SLA printed and UV laser microfabricated components which results in high yield and low variability (Fig.~\ref{fig:renders}). 

The annular emitter design used here introduces the additional constraint of requiring the emitters to be bent out of plane towards the collector grid to improve discharge performance. We accomplish this using an SLA-printed stamp and die set to hold the annulus in place while plastically deforming the tips; the emitter electrode is then lifted out of the die and placed into the final device duct. A rotating tab design and a two-layer sleeve system means that no adhesive is required to fix the electrodes in place relative to the three duct components.

The individual stages must be separated by some distance to minimize electrostatic interference (e.g., between the collector electrode of the first stage and the emitter electrode of the second), which can degrade corona discharge performance (see Fig.~\ref{fig:schematic}). Although Gilmore et al. determined a rough heuristic -- that the inter-stage gap should be about twice the inter-electrode gap in order to maximize performance~\cite{gilmore2015electrohydrodynamic}-- this was only validated using wire-to-cylinder electrode geometries. We print a variable length ducted inter-stage section which is included in the adhesiveless assembly process to investigate this further.

\subsection{Experimental Methodology}
The device under test is first fixed to a laser-cut polyetherimide (PEI) slide which can be used for either force or outlet air velocity measurement. Voltage is digitally controlled through a Spellman High Voltage SL8P supply. The resulting ion current is measured via oscilloscope across a TVS diode-protected shunt resistor. A hot wire anemometer (TSI 8465) is mounted to a motorized X/Y stage for multi-point velocity measurements. Force measurements are made using a 3D-printed test stand on a FUTEK LSB200 S-Beam load cell, with the actuator elevated to reduce possible proximity effects. Everything is fixed to a plastic optical breadboard.

\section{Experiments and Results}
\label{sec:results}
%In an effort to minimize controlled variables, all devices are designed with a two millimeter nominal inter-electrode gap (i.e., emitter tip to collector grid surface, $d$ in Fig.~\ref{fig:schematic}). As noted in the prior section, all collector grids are 25$\mu$m brass with 50 $\mu$m wire widths and 1mm wire spacing based on preliminary data. 

All experiments are performed with three devices per test condition and three trials per device. Plotted data is the overall mean (mean of each device's trial mean) with standard error of the mean (SEM) error bars. 

\subsection{Duct Size and Emitter Geometry}
We conducted experiments to investigate the hypothesis that the minimum distance between emitter tips to prevent performance degradation is related to the space charge repulsion radius $r$ predicted by the Warburg law for a specific inter-electrode gap, $r = arctan(\pi/3)d$~\cite{sigmond1986unipolar}. This corresponds to a radius of 1.6~mm for the 2~mm gap used here. Emitter tips located closer to the duct edge than this distance, or closer than twice this distance to a neighboring emitter tip (because both tips have space charge regions), are expected to show increased onset voltage and decreased ion current~\cite{moreau2008enhancing}. 

\begin{figure}
    \centering
    \includegraphics[width=\columnwidth]{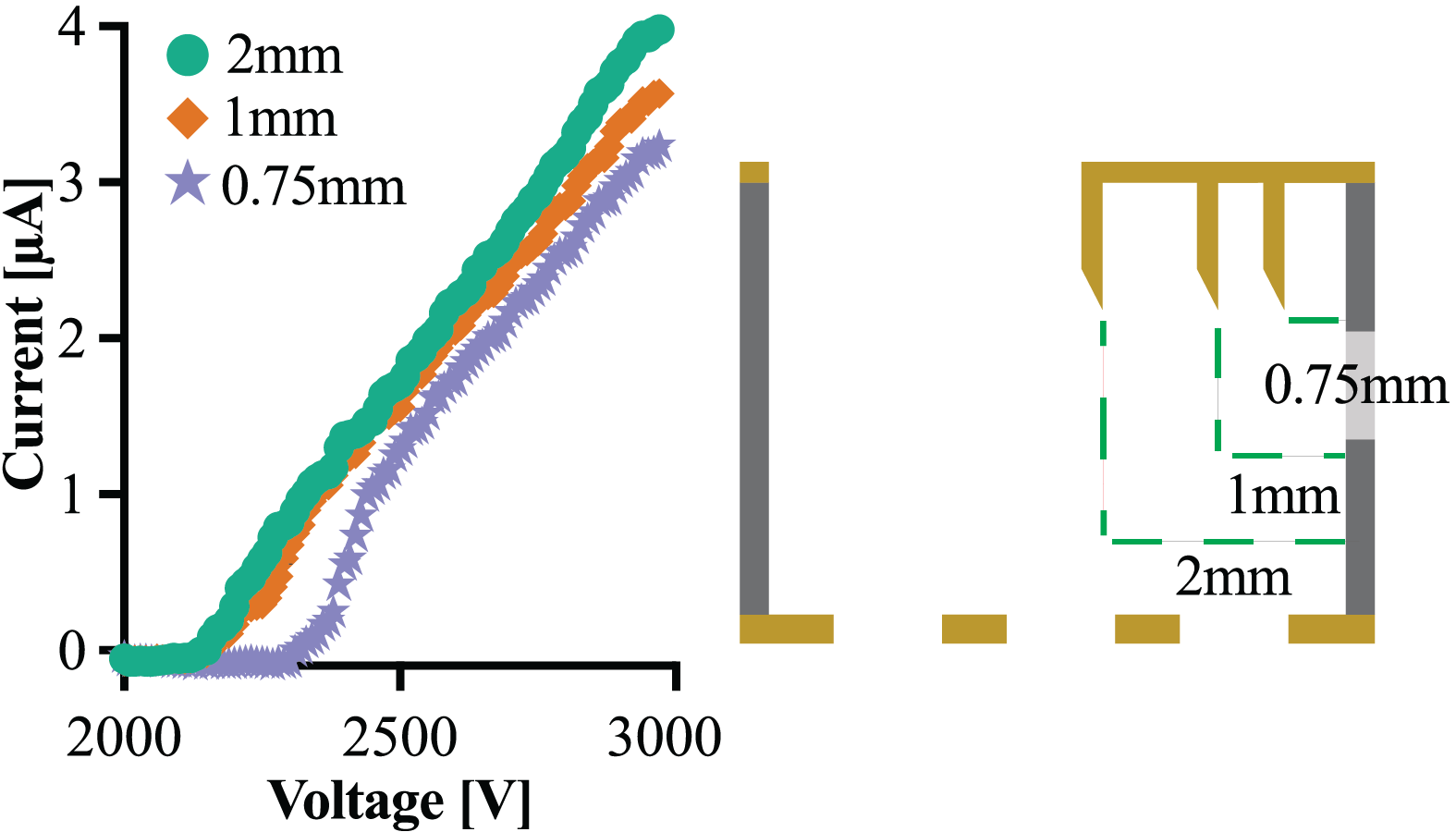}
    \caption{Current-voltage curves for 6~mm inner diameter devices with one emitter tip, with decreasing lateral distances of 2~mm, 1~mm, 0.75~mm (2.23~mm, 1.41~mm, and 1.25~mm, radially) from the annulus inner lip.}
    \label{fig:distance_from_wall}
    \vspace{-2mm}
\end{figure}

Discharge from a single emitter tip with variable distance from the annulus lip (Fig.~\ref{fig:distance_from_wall}) showed a decreased maximum current but not a significant onset voltage change for 1~mm lateral distance, which corresponds to a 1.4~mm radial distance based on the hypotenuse from the tip (bent 1~mm down) and the lip. The onset voltage increased significantly ($\approx$200~V) for the 0.75~mm (1.25~mm hypotenuse) condition. 

\begin{figure}
    \centering
    \includegraphics[width=\columnwidth]{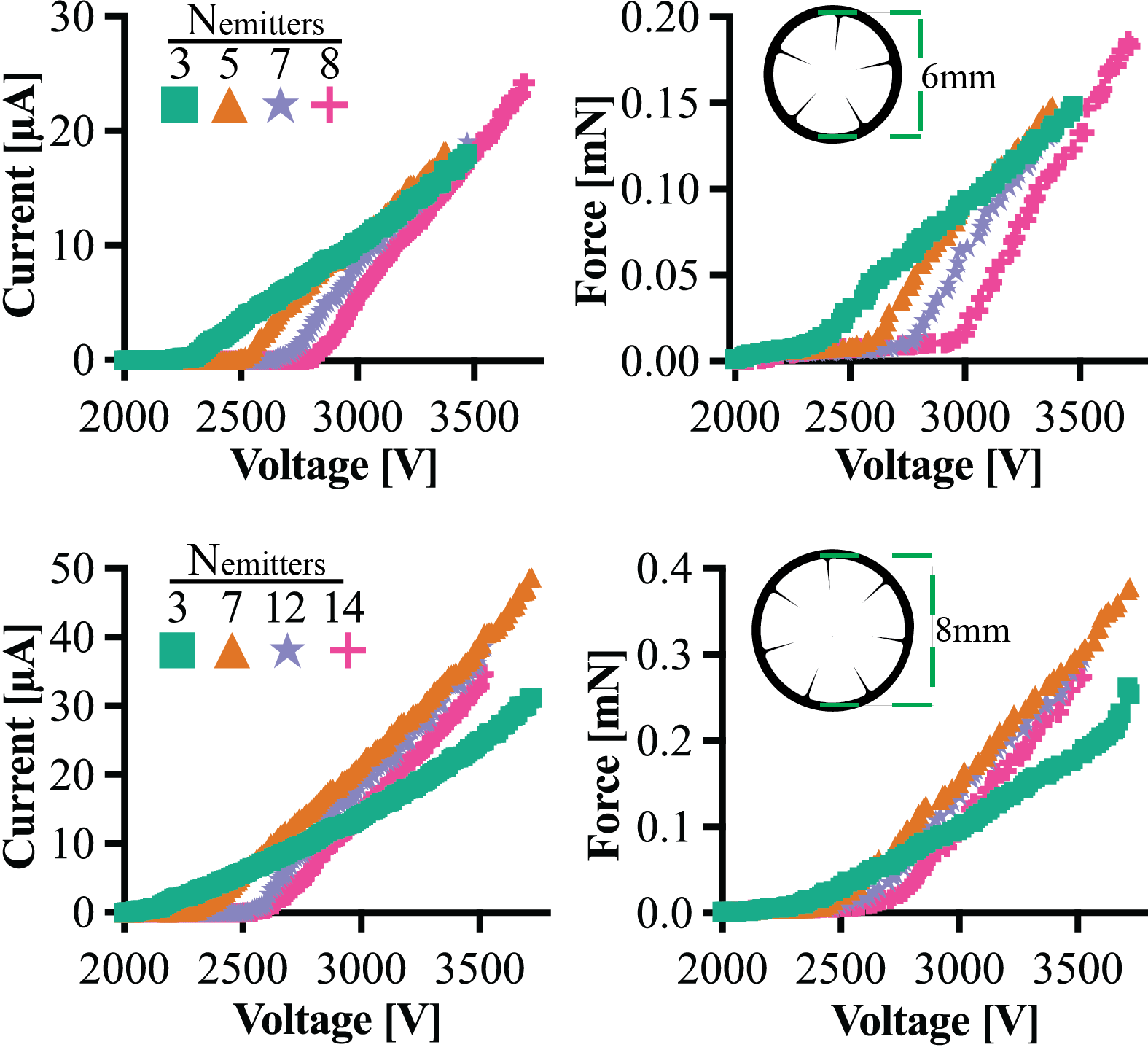}
    \caption{Current-voltage and force-voltage curves for single stage 6~mm (top) and 8~mm (bottom) inner diameter devices with different numbers of emitter tips. The 6~mm three emitter condition corresponds to a tip-to-tip distance of approximately 3.5~mm, while the five emitter distance is below the expected interference threshold from the Warburg law at only 2.3~mm.}
    \label{fig:iv_fv_6&8mm}
    \vspace{-2mm}
\end{figure}

\begin{figure}
    \centering
    \includegraphics[width=\columnwidth]{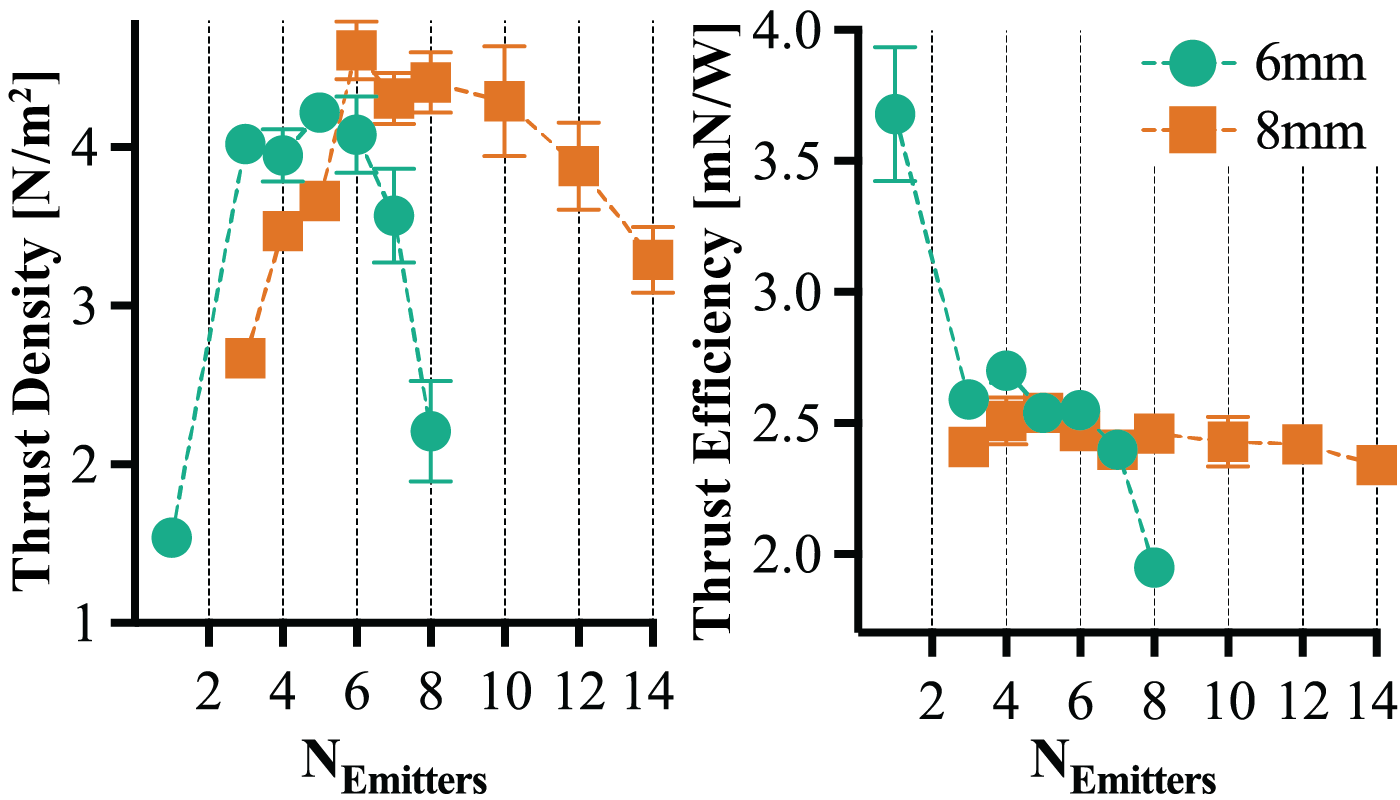}
    \caption{Thrust density and thrust efficiency versus number of emitter tips for different duct inner diameters. All data collected at 3.2~kV applied voltage; note that this is at the upper end of these devices' operating range, where the highest thrust density and lowest efficiency is expected.}
    \label{fig:emitters_thrusteta}
    %\vspace{-4mm}
\end{figure}

Experiments with multi-emitter configurations for both 6~mm and 8~mm inner diameters (Fig.~\ref{fig:iv_fv_6&8mm}) showed the expected onset voltage increase with more emitters (i.e., a decreased distance between emitters for a fixed inner diameter). There is an optimum number of tips based on operating voltage constraints; while more emitters may increase voltage, they also provide more emission points and therefore a higher potential maximum force. This trend is shown in Fig.~\ref{fig:emitters_thrusteta}, where the measured thrust density versus number of emitters reaches a local maximum before decreasing as the emitters become closer together. In the limiting case, which was not reached here, the plasma would fail to strike prior to arcing. 
% \dsd{Need to check this inter-emitter distance measurement.} The maximum thrust is achieved at 8mm inner diameter duct with six-emitters which results in a uniform distance between all emitters of approximately 3mm. These optimal dimensions were used in all subsequent experiments.   
%It is also clear that the 6mm duct diameter has higher thrust with roughly the same efficiency as the 8mm duct, implying that the space charge region does not extend through the entire inner volume of the larger duct cross section. 
%Characterization of the best performing electrode design, a 6mm duct diameter with five emitters, is shown in Fig.~\ref{fig:emitters_fviv}; at 3kV, it produces about 0.15mN of force while consuming approximately 60mW.

\subsection{Aspect Ratio}
\begin{figure}
    \centering
    \includegraphics[width=\columnwidth]{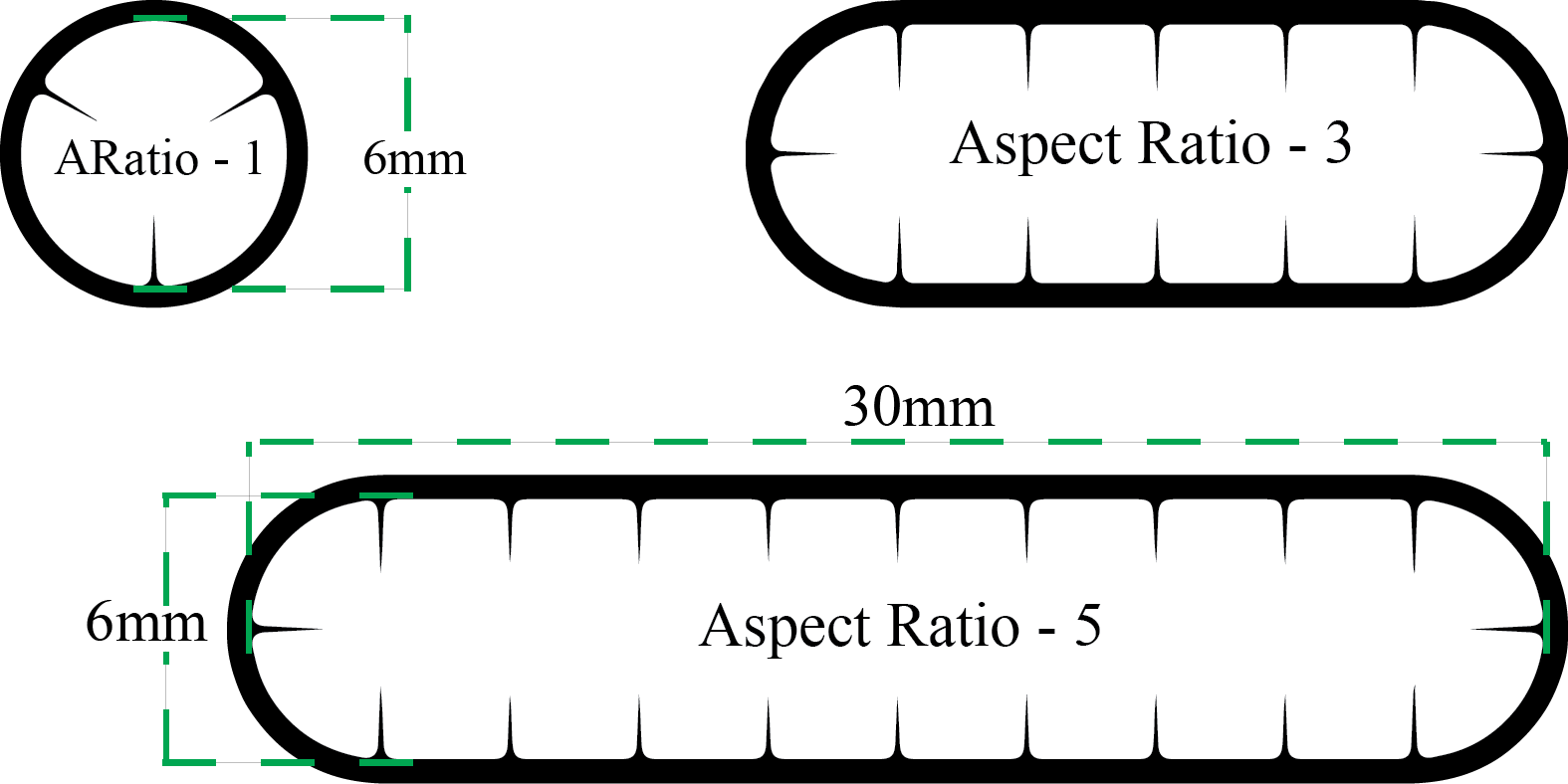}
    \caption{Schematic illustration of increasing emitter electrode aspect ratio, including a 6mm inner diameter aspect ratio one (i.e., circular) electrode (top left), an 18~mm by 6~mm aspect ratio three device (top right), and an aspect ratio five electrode with dimensions labeled (bottom).}
    \label{fig:AspectRatio}
    %\vspace{-4mm}
\end{figure}

For an annular thruster, it is evident that simply scaling the device diameter is not an effective way to increase output force; circumferential emitters are arrayed optimally based on inter-electrode gap, not diameter, and thus the center of the annulus is increasingly ``wasted'' area. We hypothesized that since the 6~mm diameter device has a nominal tip-to-tip distance across its diameter of 4~mm for the 1mm lateral placement condition (more than that necessitated by the Warburg law), elongating the duct to a higher aspect ratio (see Fig.~\ref{fig:AspectRatio}) while retaining an emitter-to-emitter distance of $\approx$3~mm along its circumference would result in a higher achievable force output without decreasing areal thrust. 

\begin{figure}
    \centering
    \includegraphics[width=\columnwidth]{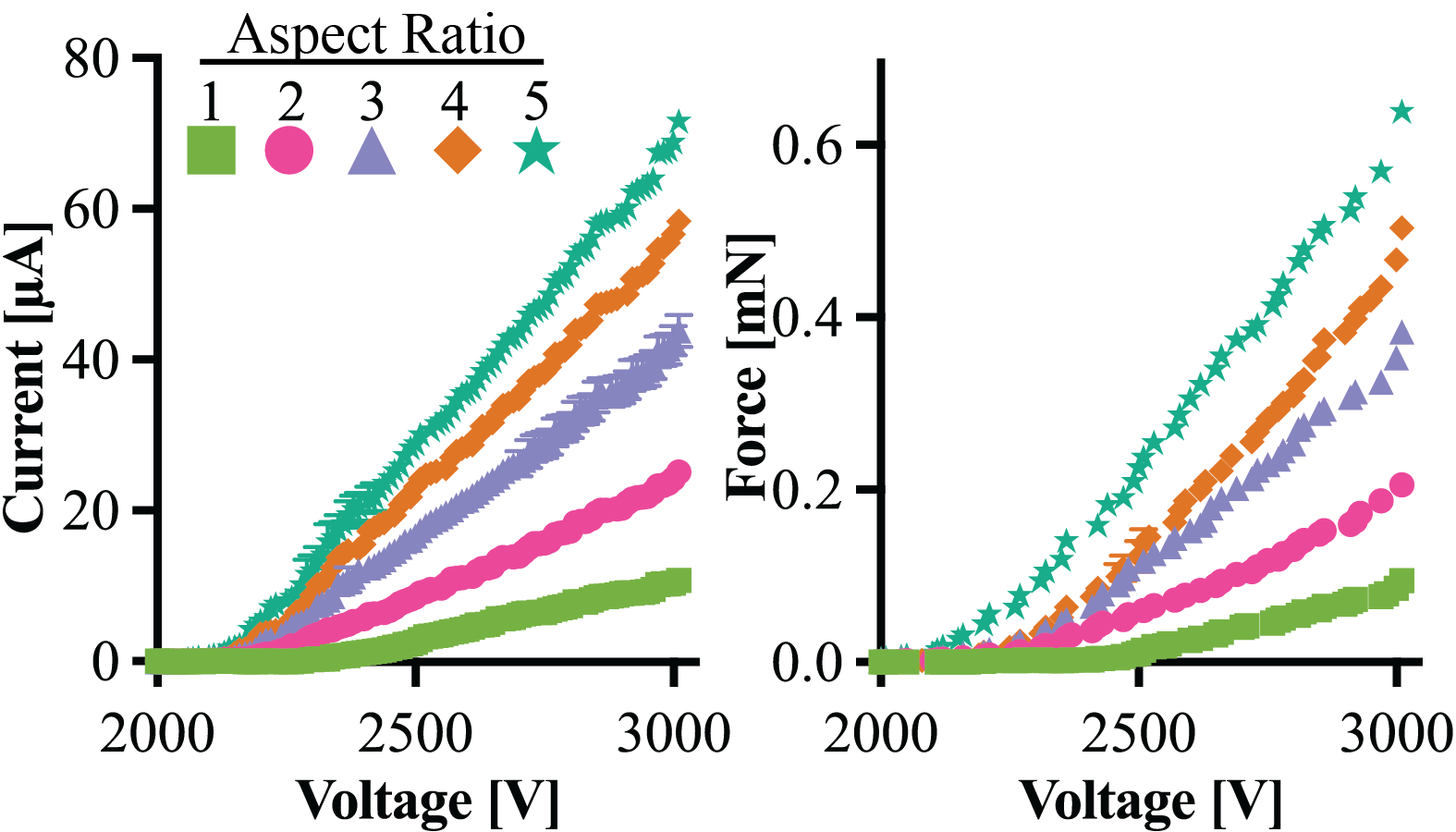}
    \caption{Current-voltage and force-voltage curves for single stage aspect ratio one through five thrusters based on the 6~mm diameter emitter-emitter-distance of 3~mm, 2~mm inter-electrode gap base design.}
    \label{fig:ar_fviv}
    \vspace{-3mm}
\end{figure}

\begin{figure}
    \centering
    \includegraphics[width=\columnwidth]{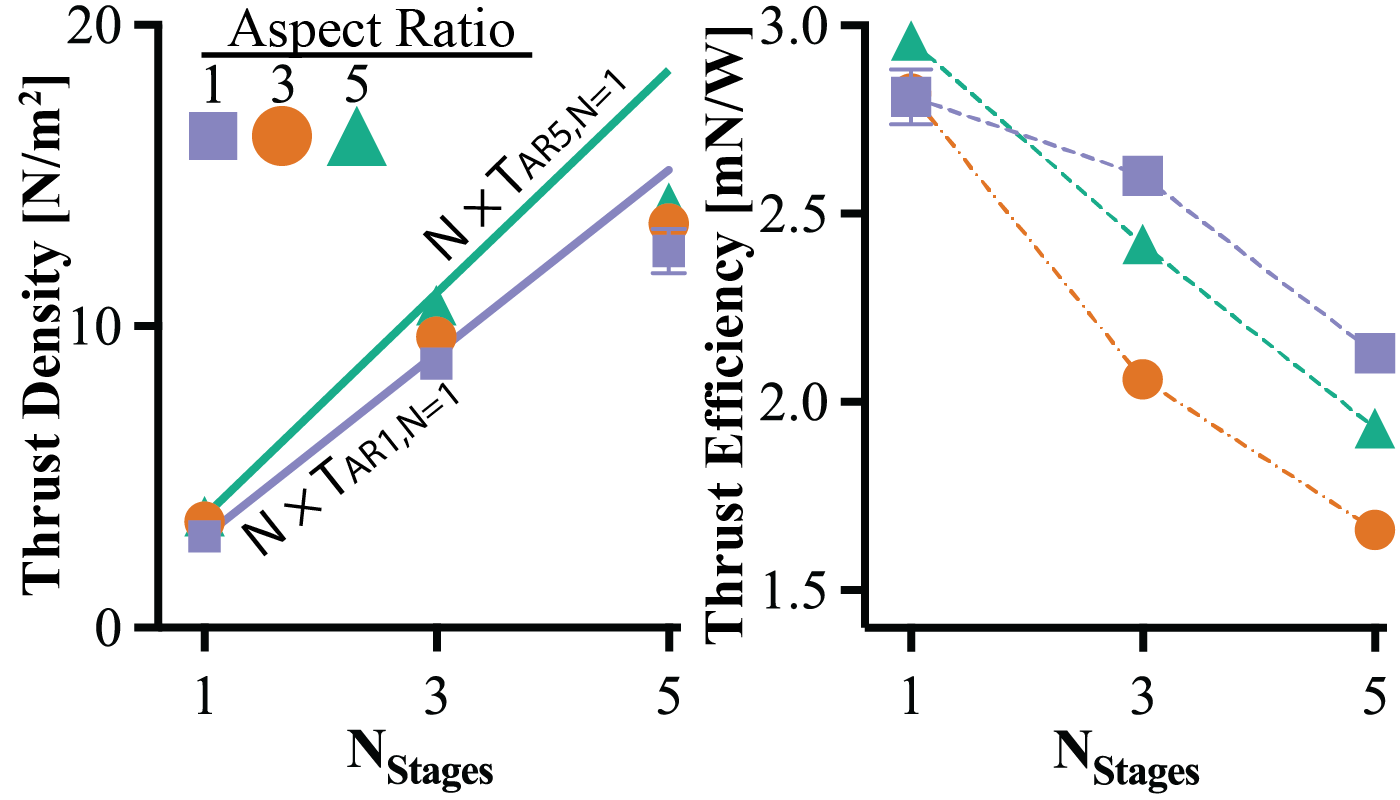}
    \caption{Thrust and thrust efficiency versus thruster aspect ratio, defined as $w/h$ where a circular duct of diameter 6~mm is an aspect ratio of one. For aspect ratios above one, emitter tips are placed to maintain approximately the same circumferential distance as in the aspect ratio one device. All data collected at 3~kV applied voltage with an inter-electrode gap of 2~mm. Solid lines represent a linear extrapolation from single-stage thrust density.}
    \label{fig:ar_thrusteta}
    \vspace{-2mm}
\end{figure}

Fig.~\ref{fig:ar_fviv} shows the current-voltage and force-voltage for single-stage devices with aspect ratios ranging from one (i.e., a circle) to five. The number of emitters for each aspect ratio is the aspect ratio multiplied by four (e.g., aspect ratio five has 20 emitters). We see a near linear increase in total output force with increasing aspect ratio, indicating that our device scaling methodology was successful. Fig.~\ref{fig:ar_thrusteta} shows that the thrust and thrust efficiency remain approximately constant as the aspect ratio increases from one to five for a single stage. The data for stage counts above one are discussed later.

\subsection{Inter-stage Length}
One of the challenges with multi-stage devices is managing the electrostatic interactions between active electrodes in neighboring stages (e.g., the collector of stage one and the emitter of stage two). We investigated the hypothesis that our device structure would allow us to decrease the interstage distance ratio below $2d$, a value experimentally determined in prior work by Gilmore et al.~\cite{gilmore2015electrohydrodynamic} and Drew et al.~\cite{drew2021high}, because the annular bent emitters effectively shields the tips from the preceding stage. 
% Decreasing the inter-stage distance is desirable because it translates to an increase in potential thrust-to-weight ratio and power density.

\begin{figure}
    \centering
    \includegraphics{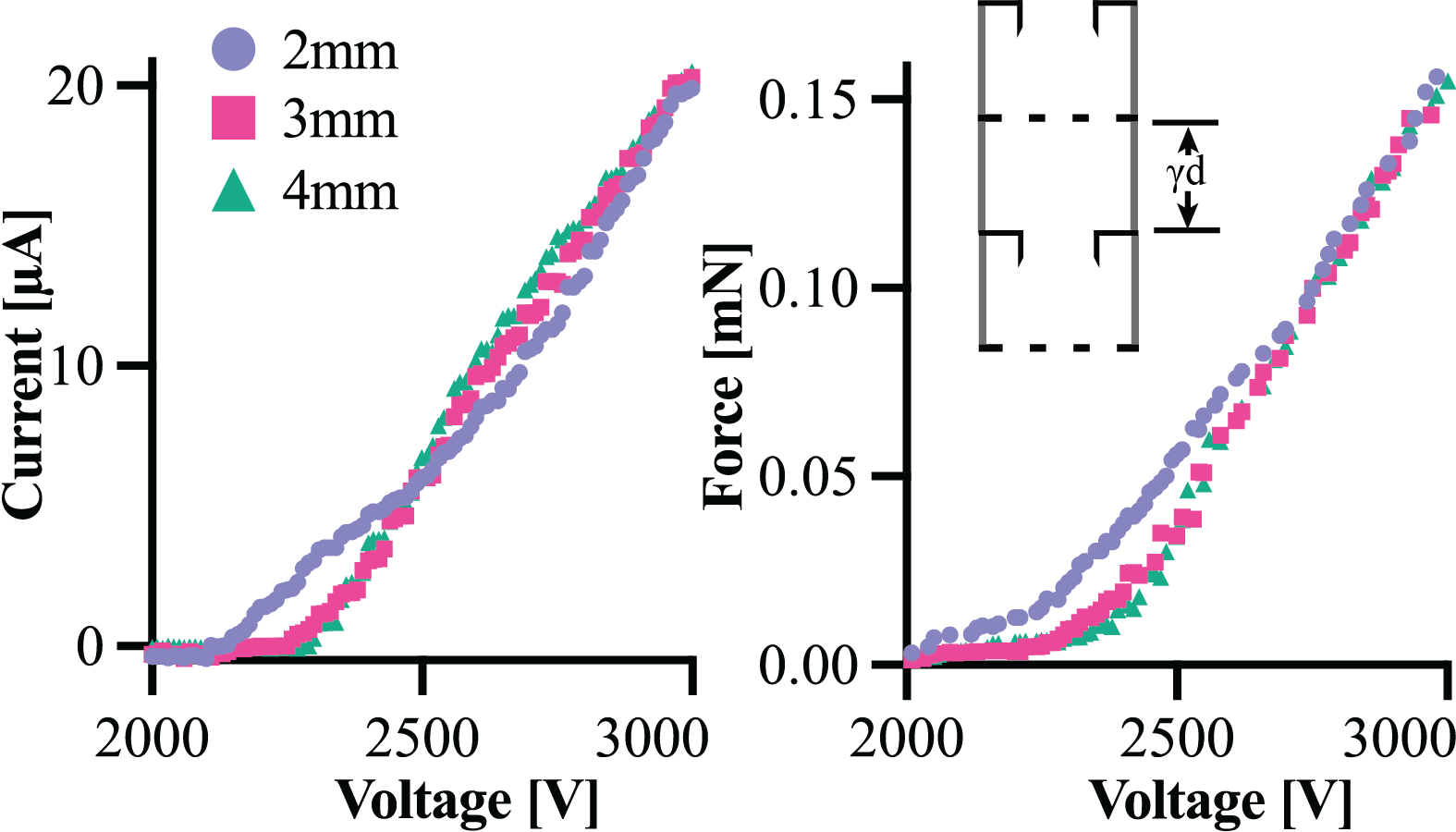}
    \caption{Results for variable inter-stage distances of 2~mm, 3~mm, and 4~mm, corresponding to $\gamma$ values of 1, 1.5, and 2.}
    \label{fig:interstage}
    \vspace{-5mm}
\end{figure}

Fig.~\ref {fig:interstage} shows results for the inter-stage distances of $2d$, $1.5d$, and $d$, which exhibit near identical performance in a two-stage device above about 2750~V. Below $d$, arcing along the duct surface between the outer edges of the emitter and collector was the limiting factor. Interestingly, onset voltage appears to \textit{decrease} for the 2~mm ($1d$) condition. Some devices would arc between the collector of stage one and the bent corners of the stage two emitters prior to discharge (data from those ``failed'' devices not shown). The decreased onset voltage and interesting shape of the IV curve could possibly be attributed to reverse corona discharge from the emitter corners to the prior collector, with some percentage of that plasma's ejected ions actually drifting instead towards the ``correct'' collector. To remove this unpredictable element we performed all subsequent experiments at an inter-stage gap of 3~mm ($1.5d$), still under the inter-stage distance ratio of $2d$ previously determined by Gilmore et al.~\cite{gilmore2015electrohydrodynamic}.

\subsection{Stage Count}
Prior experimental work on centimeter-scale multi-stage ducted EAD devices exhibited sub-linear scaling of output force with number of stages, although characterization was preliminary and device geometries were not explored in depth~\cite{drew2021high}. Our hypothesis was that the careful process control we took for developing low-friction duct wall surfaces would reduce inter-stage losses. Conversely, preliminary modeling showed that the flow would fail to fully develop in the duct lengths shown here at the low Reynolds numbers achievable by these devices, especially given the turbulence induced by passage through the collector grids. We therefore expected no efficiency benefit as may have been predicted by recent analytical modeling efforts~\cite{gomez2023model}. 

Fig.~\ref{fig:ar_thrusteta} shows the thrust efficiency acutely drops with an increase in stages. A possible explanation is that efficiency losses are largely due to fluid flow damping in the inter-stage region (see Fig.~\ref{fig:schematic}) due to turbulence generated from the collector grid, with minor losses from the surface shear stress that is compensated for by additional emitters resulting in nearly the same efficiency for a single stage device at different aspect ratios.
This loss causes increased deviation from the density that would be predicted for multiple stages given a linear fit (see the \textit{N $\times$ T} lines in Fig.~\ref{fig:ar_thrusteta}).

\begin{figure}
    \centering
    \includegraphics[width=\columnwidth]{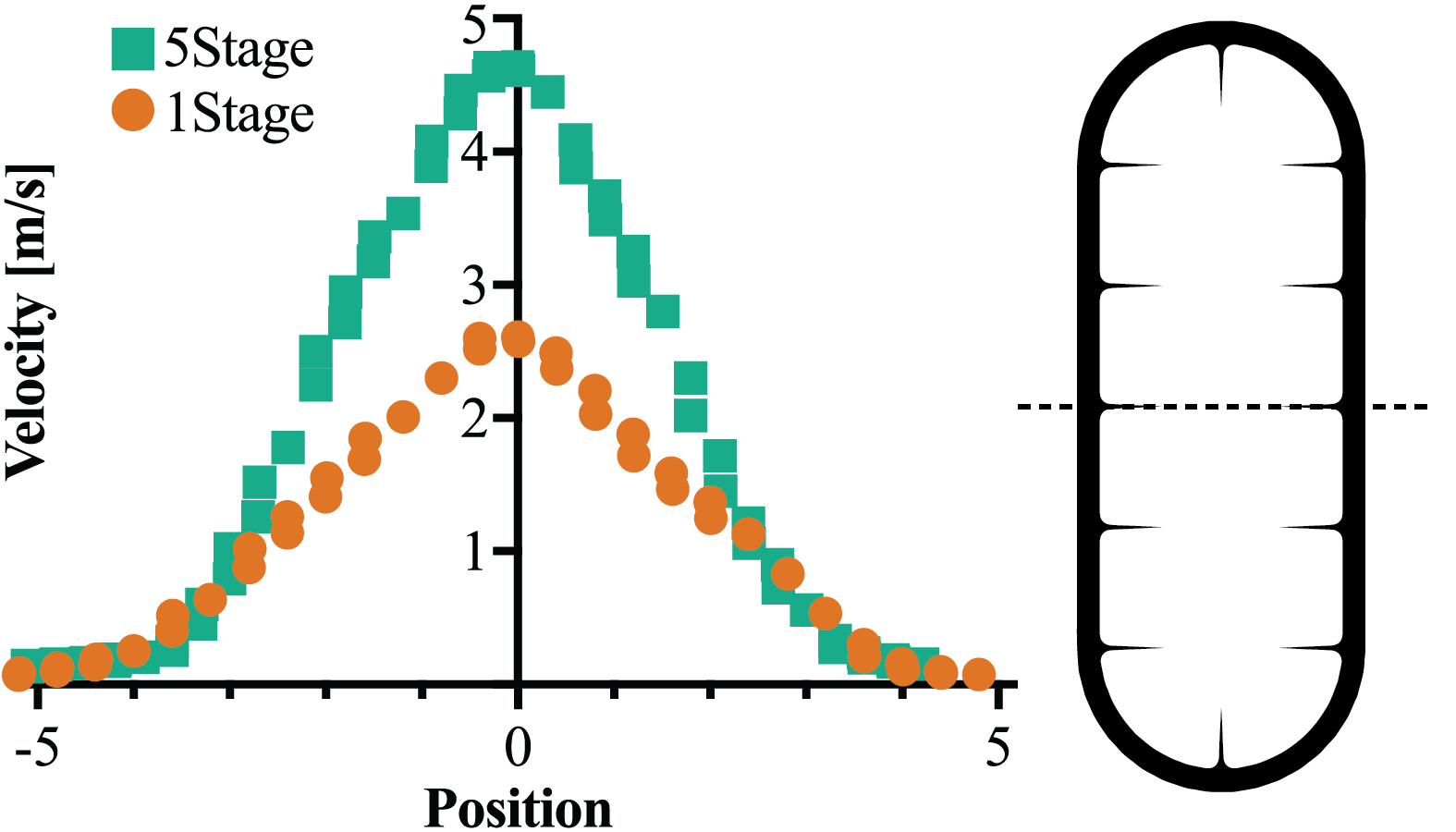}
    \caption{Two dimensional air velocity data measured at the outlet of one-stage and five-stage aspect ratio three thrusters.}
    \label{fig:anemometer}
    \vspace{-5mm}
\end{figure}

Fig.~\ref{fig:anemometer} shows outlet air velocity data measured from one-stage and five-stage aspect ratio three devices via a hot wire anemometer. The characteristic convergent jet plume expected from a ducted actuator is evident in both cases, indicating that models for fluid dynamic/aerodynamic efficiency gains hypothesized for circular ducted actuators should extend to high aspect ratio devices. Taking the 6mm axis as the chord length, the Reynolds number for the five-stage device is approximately 2000, falling in the transition region from laminar to turbulent flow where analytical models often fail.

\subsection{Distributed Propulsion}
\begin{figure}
    \centering
    \includegraphics{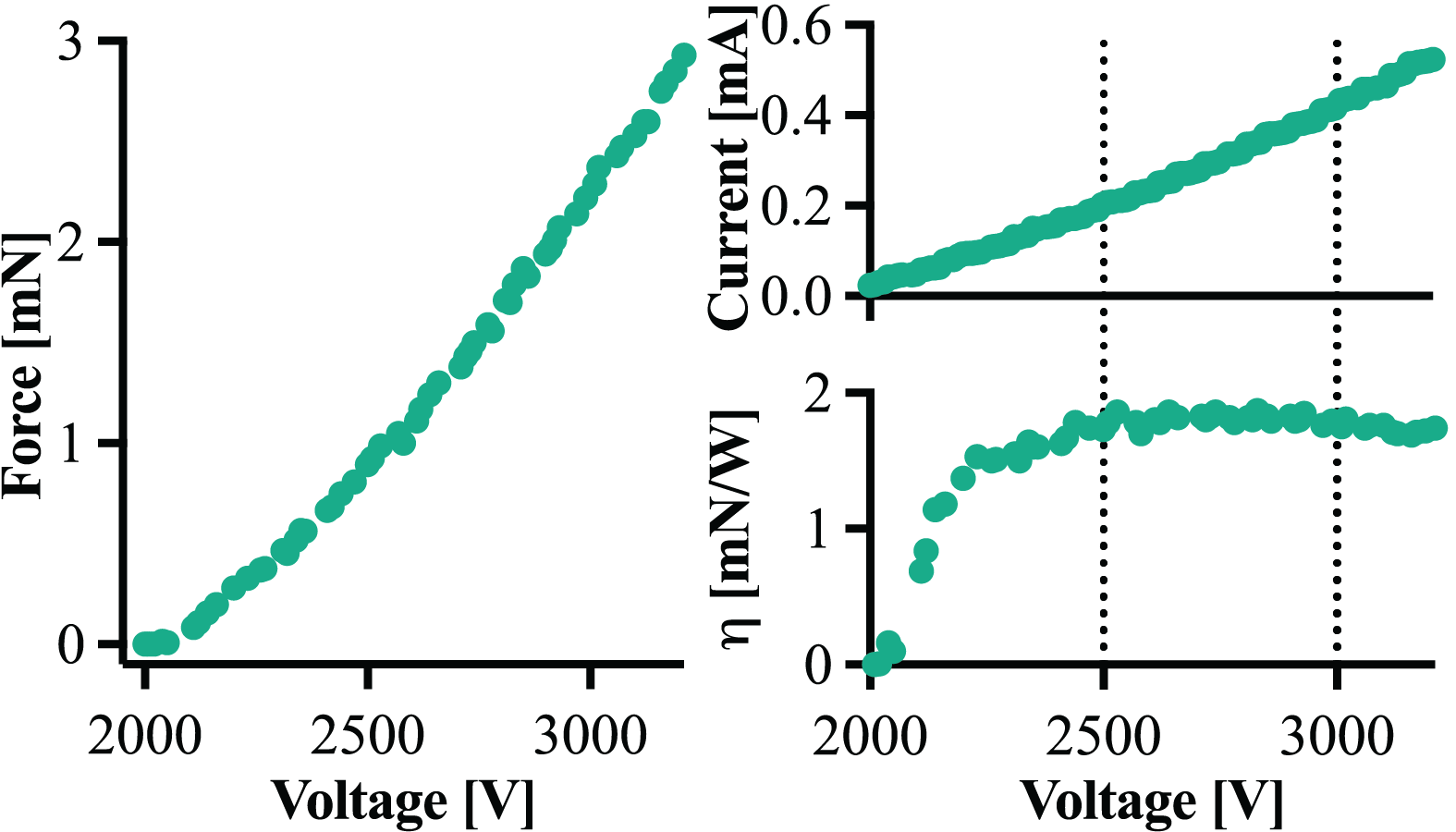}
    \caption{Characterization of a five-stage aspect ratio five device operated up to approximately 3.3~kV.}
    \label{fig:best}
    \vspace{-1mm}
\end{figure}

The peak performance achieved from our five-stage aspect ratio five device is a thrust of 17.93N/m$^2$ (3.09~mN output force with 172.27~mm$^2$ inner area) at 3.28~kV, with an efficiency of 1.86~mN/W (Fig.~\ref{fig:best}). Compared to state-of-the-art flapping wing platforms, like the DEA-based design presented by Ren et al.~\cite{ren_high-lift_2022} with a calculated thrust density of $\approx$12~N/m$^2$ and a thrust efficiency of 10~mN/W (and up to 20~mN/W for a higher voltage design), this is a significantly higher thrust density and a significantly lower efficiency. It is important to note that the efficiency values derived for flapping wing- and propeller-based fliers include the lift generated by the wings/rotors. Assuming that a lift-to-drag ratio between 5 and 10 is possible for a centimeter-scale MAV~\cite{wood2005design}, our achievable efficiency actually looks promising for a non-VTOL platform.

\begin{figure}
    \centering
    \includegraphics[width=\columnwidth]{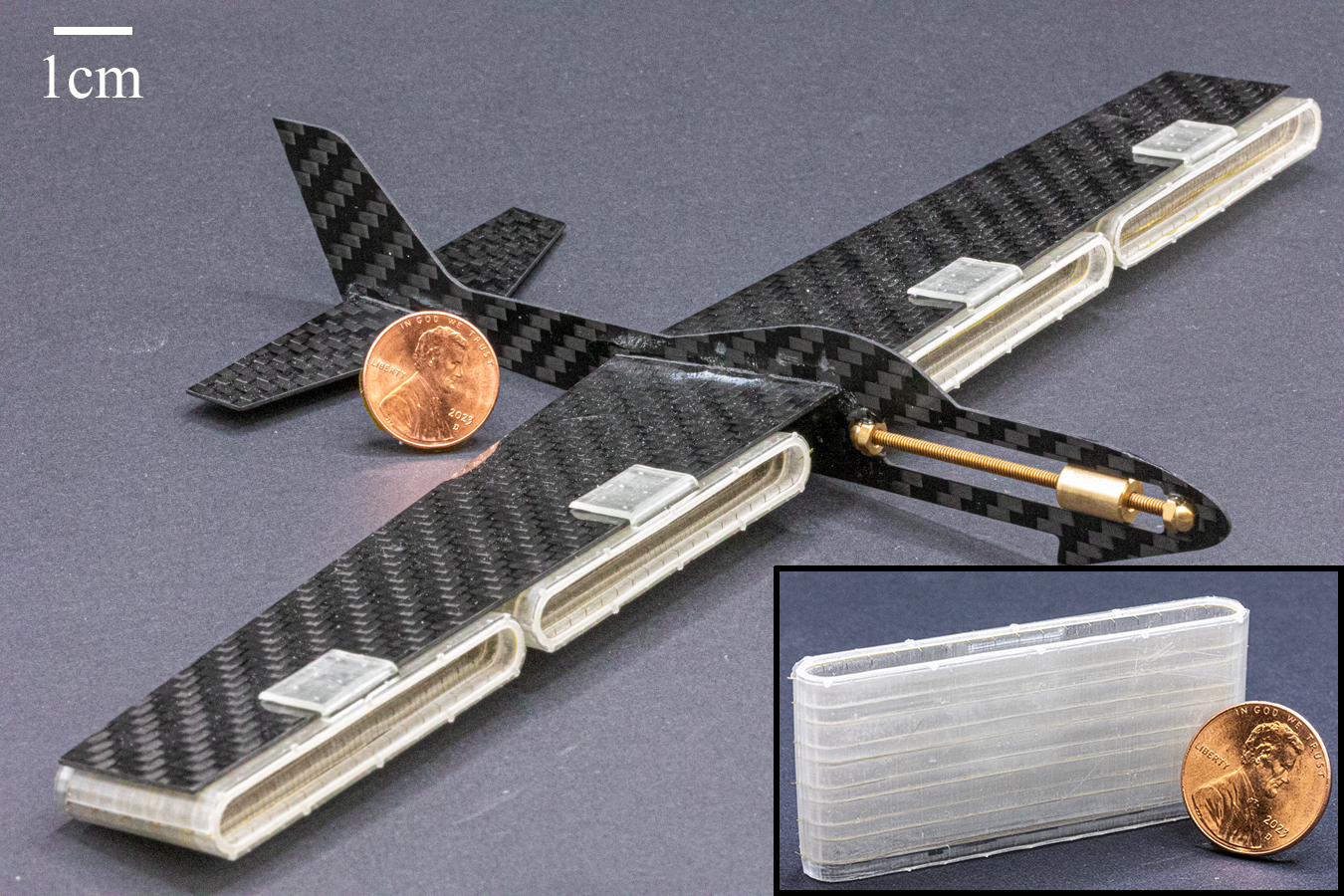}
    \caption{Proof of concept for a commercial MAV-scale carbon fiber glider with five-stage, aspect ratio nine thrusters mounted underwing along the leading edge. The existing ballast mass seen in the nose of this commercial platform could be replaced by power and control electronics.}
    \label{fig:glider}
    \vspace{-4mm}
\end{figure}

We hypothesize that the narrow form factor afforded by these high aspect ratio actuators lends itself to distributed propulsion on fixed-wing platforms, where it is important to minimize the impact of the actuator on the aerodynamic performance of the airfoil (i.e., minimize viscous drag~\cite{xu2018flight}). Fig.~\ref{fig:glider} shows five-stage aspect ratio nine devices integrated under the wing of a commercial MAV-scale glider as a proof-of-concept. The four aspect ratio nine thrusters are expected to produce about 22~mN of combined force (17.58~N/m$^2$, 316.27~mm$^2$ thruster area); with sufficient lift from the wings and modifications to improve thrust-to-weight ratio, that could be sufficient for short power autonomous flight. 

\section{Conclusion and Future Work}
\label{sec:conclusion}
We show that serial integration of multiple acceleration stages in a ducted electroaerodynamic actuator increases thrust density to levels exceeding existing alternatives, and elongating to high aspect ratios increases output force while providing a suitable form factor for distributed propulsion on fixed-wing micro air vehicles. We demonstrate a novel device design with among the highest thrust density and efficiency for an EAD actuator ever measured at this scale ($\approx$18 N/m$^2$ and $\approx$2 mN/W, respectively). A new adhesiveless assembly process results in reliable devices as evidenced by the relatively low variation in the experimental data, traditionally a challenge at this scale. 

A major limitation of this work is that the duct design was optimized for rapid experimentation with discrete devices, which resulted in a low thrust-to-weight ratio unsuitable for flight. Replacing the SLA-printed ducts with either UV laser micromachined Kapton tubing or two-photon-polymerization printed thin wall structures would drastically reduce mass; we estimate that a thrust-to-weight ratio of over 2.5 is achievable without significant redesign effort.

Even though the demonstrated actuators are high performance relative to other centimeter-scale EAD devices, the thrust efficiency is still too low for use in a power autonomous MAV. In future work, higher stage counts, necessary to increase Reynolds number, could yield efficiency gains through either fluid dynamic (e.g., pressure driven flow out an exhaust nozzle) or aerodynamic (e.g., lift generated by an airfoil intake lip) mechanisms. This may require a redesign of the collector electrode to allow for fully developed flow without grid-induced turbulence.

\section*{ACKNOWLEDGMENT}
The authors would like to thank Rebecca Miles and Nichols Crawford Taylor for their help designing the experimental setup and the University of Utah Nanofab staff for their help developing a repeatable UV-laser process.

\bibliographystyle{IEEEtran}
\bibliography{root}  % .bib

\begin{thebibliography}{10}
\providecommand{\url}[1]{#1}
\csname url@rmstyle\endcsname
\providecommand{\newblock}{\relax}
\providecommand{\bibinfo}[2]{#2}
\providecommand\BIBentrySTDinterwordspacing{\spaceskip=0pt\relax}
\providecommand\BIBentryALTinterwordstretchfactor{4}
\providecommand\BIBentryALTinterwordspacing{\spaceskip=\fontdimen2\font plus
\BIBentryALTinterwordstretchfactor\fontdimen3\font minus
  \fontdimen4\font\relax}
\providecommand\BIBforeignlanguage[2]{{%
\expandafter\ifx\csname l@#1\endcsname\relax
\typeout{** WARNING: IEEEtran.bst: No hyphenation pattern has been}%
\typeout{** loaded for the language `#1'. Using the pattern for}%
\typeout{** the default language instead.}%
\else
\language=\csname l@#1\endcsname
\fi
#2}}

\bibitem{floreano2015science}
D.~Floreano and R.~J. Wood, ``Science, technology and the future of small
  autonomous drones,'' \emph{Nature}, vol. 521, no. 7553, pp. 460--466, 2015.

\bibitem{dorigo2020reflections}
M.~Dorigo, G.~Theraulaz, and V.~Trianni, ``Reflections on the future of swarm
  robotics,'' \emph{Science Robotics}, vol.~5, no.~49, p. eabe4385, 2020.

\bibitem{ward2017bibliometric}
T.~A. Ward, C.~J. Fearday, E.~Salami, and N.~Binti~Soin, ``A bibliometric
  review of progress in micro air vehicle research,'' \emph{International
  Journal of Micro Air Vehicles}, vol.~9, no.~2, pp. 146--165, 2017.

\bibitem{talwekar2022towards}
Y.~P. Talwekar, A.~Adie, V.~Iyer, and S.~B. Fuller, ``Towards sensor autonomy
  in sub-gram flying insect robots: A lightweight and power-efficient avionics
  system,'' in \emph{2022 International Conference on Robotics and Automation
  (ICRA)}.\hskip 1em plus 0.5em minus 0.4em\relax IEEE, 2022, pp. 9675--9681.

\bibitem{james2018liftoff}
J.~James, V.~Iyer, Y.~Chukewad, S.~Gollakota, and S.~B. Fuller, ``Liftoff of a
  190 mg laser-powered aerial vehicle: The lightest wireless robot to fly,'' in
  \emph{2018 IEEE International Conference on Robotics and Automation
  (ICRA)}.\hskip 1em plus 0.5em minus 0.4em\relax IEEE, 2018, pp. 3587--3594.

\bibitem{zhou2022swarm}
X.~Zhou, X.~Wen, Z.~Wang, Y.~Gao, H.~Li, Q.~Wang, T.~Yang, H.~Lu, Y.~Cao,
  C.~Xu, \emph{et~al.}, ``Swarm of micro flying robots in the wild,''
  \emph{Science Robotics}, vol.~7, no.~66, p. eabm5954, 2022.

\bibitem{mulgaonkar2014power}
Y.~Mulgaonkar, M.~Whitzer, B.~Morgan, C.~M. Kroninger, A.~M. Harrington, and
  V.~Kumar, ``Power and weight considerations in small, agile quadrotors,'' in
  \emph{Micro-and Nanotechnology Sensors, Systems, and Applications VI}, vol.
  9083.\hskip 1em plus 0.5em minus 0.4em\relax SPIE, 2014, pp. 376--391.

\bibitem{schaffer2021drone}
B.~Sch{\"a}ffer, R.~Pieren, K.~Heutschi, J.~M. Wunderli, and S.~Becker, ``Drone
  noise emission characteristics and noise effects on humans—a systematic
  review,'' \emph{International Journal of Environmental Research and Public
  Health}, vol.~18, no.~11, p. 5940, 2021.

\bibitem{wood2012progress}
R.~J. Wood, B.~Finio, M.~Karpelson, K.~Ma, N.~O. P{\'e}rez-Arancibia, P.~S.
  Sreetharan, H.~Tanaka, and J.~P. Whitney, ``Progress on ‘pico’air
  vehicles,'' \emph{The International Journal of Robotics Research}, vol.~31,
  no.~11, pp. 1292--1302, 2012.

\bibitem{ren_high-lift_2022}
Z.~Ren, S.~Kim, X.~Ji, W.~Zhu, F.~Niroui, J.~Kong, and Y.~Chen,
  ``\BIBforeignlanguage{en}{A {High}-{Lift} {Micro}-{Aerial}-{Robot} {Powered}
  by {Low}-{Voltage} and {Long}-{Endurance} {Dielectric} {Elastomer}
  {Actuators}},'' \emph{\BIBforeignlanguage{en}{Advanced Materials}}, vol.~34,
  no.~7, p. 2106757, 2022.

\bibitem{xu2018flight}
H.~Xu, Y.~He, K.~L. Strobel, C.~K. Gilmore, S.~P. Kelley, C.~C. Hennick,
  T.~Sebastian, M.~R. Woolston, D.~J. Perreault, and S.~R. Barrett, ``Flight of
  an aeroplane with solid-state propulsion,'' \emph{Nature}, vol. 563, no.
  7732, pp. 532--535, 2018.

\bibitem{drew2018toward}
D.~S. Drew, N.~O. Lambert, C.~B. Schindler, and K.~S. Pister, ``Toward
  controlled flight of the ionocraft: a flying microrobot using
  electrohydrodynamic thrust with onboard sensing and no moving parts,''
  \emph{IEEE Robotics and Automation Letters}, vol.~3, no.~4, pp. 2807--2813,
  2018.

\bibitem{pekker2011model}
L.~Pekker and M.~Young, ``Model of ideal electrohydrodynamic thruster,''
  \emph{journal of propulsion and power}, vol.~27, no.~4, pp. 786--792, 2011.

\bibitem{masuyama2013performance}
K.~Masuyama and S.~R. Barrett, ``On the performance of electrohydrodynamic
  propulsion,'' \emph{Proceedings of the Royal Society A: Mathematical,
  Physical and Engineering Sciences}, vol. 469, no. 2154, p. 20120623, 2013.

\bibitem{gilmore2015electrohydrodynamic}
C.~K. Gilmore and S.~R. Barrett, ``Electrohydrodynamic thrust density using
  positive corona-induced ionic winds for in-atmosphere propulsion,''
  \emph{Proceedings of the Royal Society A: Mathematical, Physical and
  Engineering Sciences}, vol. 471, no. 2175, p. 20140912, 2015.

\bibitem{drew2021high}
D.~S. Drew and S.~Follmer, ``High force density multi-stage electrohydrodynamic
  jets using folded laser microfabricated electrodes,'' in \emph{2021 21st
  International Conference on Solid-State Sensors, Actuators and Microsystems
  (Transducers)}.\hskip 1em plus 0.5em minus 0.4em\relax IEEE, 2021, pp.
  54--57.

\bibitem{gomez2023model}
N.~Gomez-Vega, A.~Brown, H.~Xu, and S.~R. Barrett, ``Model of multistaged
  ducted thrusters for high-thrust-density electroaerodynamic propulsion,''
  \emph{AIAA Journal}, vol.~61, no.~2, pp. 767--779, 2023.

\bibitem{drew_first_2017}
D.~S. Drew and K.~S.~J. Pister, ``First takeoff of a flying microrobot with no
  moving parts,'' in \emph{2017 {International} {Conference} on {Manipulation},
  {Automation} and {Robotics} at {Small} {Scales} ({MARSS})}, July 2017, pp.
  1--5.

\bibitem{prasad_laser-microfabricated_2020}
H.~K.~H. Prasad, R.~S. Vaddi, Y.~M. Chukewad, E.~Dedic, I.~Novosselov, and
  S.~B. Fuller, ``\BIBforeignlanguage{en}{A laser-microfabricated
  electrohydrodynamic thruster for centimeter-scale aerial robots},''
  \emph{\BIBforeignlanguage{en}{PLOS ONE}}, vol.~15, no.~4, p. e0231362, Apr.
  2020.

\bibitem{zhang_passive_2022}
H.~Zhang, J.~Leng, Z.~Liu, M.~Qi, and X.~Yan, ``\BIBforeignlanguage{en}{Passive
  attitude stabilization of ionic-wind-powered micro air vehicles},''
  \emph{\BIBforeignlanguage{en}{Chinese Journal of Aeronautics}}, p.
  S1000936122003107, Dec. 2022.

\bibitem{zhang_centimeter-scale_2022}
H.~Zhang, J.~Leng, D.~Liu, W.~Zhan, R.~Yun, Z.~Liu, M.~Qi, and X.~Yan, ``A
  {Centimeter}-{Scale} {Electrohydrodynamic} {Multi}-{Modal} {Robot} {Capable}
  of {Rolling}, {Hopping}, and {Taking} {Off},'' \emph{IEEE Robotics and
  Automation Letters}, vol.~7, no.~4, pp. 11\,791--11\,798, Oct. 2022,
  conference Name: IEEE Robotics and Automation Letters.

\bibitem{kim2010velocity}
C.~Kim, D.~Park, K.~Noh, and J.~Hwang, ``Velocity and energy conversion
  efficiency characteristics of ionic wind generator in a multistage
  configuration,'' \emph{Journal of Electrostatics}, vol.~68, no.~1, pp.
  36--41, 2010.

\bibitem{moreau2008enhancing}
E.~Moreau and G.~Touchard, ``Enhancing the mechanical efficiency of electric
  wind in corona discharges,'' \emph{Journal of Electrostatics}, vol.~66, no.
  1-2, pp. 39--44, 2008.

\bibitem{pereira2008hover}
J.~L. Pereira, \emph{Hover and wind-tunnel testing of shrouded rotors for
  improved micro air vehicle design}.\hskip 1em plus 0.5em minus 0.4em\relax
  University of Maryland, College Park, 2008.

\bibitem{hrishikeshavan2012design}
V.~Hrishikeshavan, J.~Black, and I.~Chopra, ``Design and testing of a quad
  shrouded rotor micro air vehicle in hover,'' in \emph{53rd
  aiaa/aSME/aSCE/aHS/aSC Structures, Structural Dynamics and Materials
  Conference 20th aiaa/aSME/aHS adaptive Structures Conference 14th aiaa},
  2012, p. 1720.

\bibitem{drew2017geometric}
D.~S. Drew and K.~S. Pister, ``Geometric optimization of microfabricated
  silicon electrodes for corona discharge-based electrohydrodynamic
  thrusters,'' \emph{Micromachines}, vol.~8, no.~5, p. 141, 2017.

\bibitem{sigmond1986unipolar}
R.~Sigmond, ``The unipolar corona space charge flow problem,'' \emph{Journal of
  Electrostatics}, vol.~18, no.~3, pp. 249--272, 1986.

\bibitem{wood2005design}
R.~Wood, S.~Avadhanula, E.~Steltz, M.~Seeman, J.~Entwistle, A.~Bachrach,
  G.~Barrows, S.~Sanders, and R.~Fearing, ``Design, fabrication and initial
  results of a 2g autonomous glider,'' in \emph{31st Annual Conference of IEEE
  Industrial Electronics Society, 2005. IECON 2005.}\hskip 1em plus 0.5em minus
  0.4em\relax IEEE, 2005, pp. 8--pp.

\end{thebibliography}

\end{document}